%% file: main.tex
\documentclass[journal]{IEEEtran}

\usepackage{cite}
\usepackage[pdftex]{graphicx}
\usepackage{amsmath}
\usepackage{array}
\usepackage[caption=false,font=footnotesize]{subfig}
\usepackage{fixltx2e}
\usepackage{url,hyperref}
\usepackage{multirow}
\usepackage[table]{xcolor}
\usepackage{color}
\usepackage{caption}
\usepackage{booktabs}
\usepackage[justification=centering]{caption} 
\usepackage{epsfig,stfloats,amssymb,makecell,flushend}
\usepackage[ruled]{algorithm2e}
\usepackage{ragged2e}
\SetAlgoSkip{}

\hypersetup{
    colorlinks=true
}

\definecolor{ccp}{RGB}{0,176,240}
\definecolor{msp}{RGB}{0,176,80}
\definecolor{map}{RGB}{181,100,20}

\begin{document}

\title{Resilient Multimodal Industrial Surface Defect Detection with Uncertain Sensors Availability}

\author{Shuai~Jiang,
        Yunfeng~Ma,
        Jingyu~Zhou,
        Yuan~Bian,
        Yaonan~Wang,
        and~Min~Liu 
\thanks{This work was supported in part by the National Key Research and Development Program of China under Grant 2022YFB3303800, in part by the Natural Science Foundation of Hunan Province under Grant 2024JJ3013, in part by the National Natural Science Foundation of China under Grant 62425305, in part by the Science and Technology Innovation Program of Hunan Province under Grant 2023RC1048. (Corresponding author: Min Liu.)}
\thanks{The authors are with the School of Artificial Intelligence and Robotics and the National Engineering Research Center for Robot Visual Perception and Control Technology, Hunan University, Changsha, Hunan 410082, China.}}

\markboth{Journal of \LaTeX\ Class Files,~Vol.~14, No.~8, August~2015}%
{Shell \MakeLowercase{\textit{et al.}}: Bare Demo of IEEEtran.cls for IEEE Journals}

\maketitle

\input{Chapters/0_Abs}


\IEEEpeerreviewmaketitle

\input{Chapters/1_Intro}
\input{Chapters/2_Rel}
\input{Chapters/3_Met}
\input{Chapters/4_Exps}
\input{Chapters/5_Conc}

\bibliographystyle{ieeetr}
\bibliography{ref_abbr}

\end{document}

%% file: Chapters/0_Abs.tex
\begin{abstract}
Multimodal industrial surface defect detection (MISDD) aims to identify and locate defect in industrial products by fusing RGB and 3D modalities. This article focuses on modality-missing problems caused by uncertain sensors availability in MISDD. In this context, the fusion of multiple modalities encounters several troubles, including learning mode transformation and information vacancy. To this end, we first propose cross-modal prompt learning, which includes: i) the cross-modal consistency prompt serves the establishment of information consistency of dual visual modalities; ii) the modality-specific prompt is inserted to adapt different input patterns; iii) the missing-aware prompt is attached to compensate for the information vacancy caused by dynamic modalities-missing. In addition, we propose symmetric contrastive learning, which utilizes text modality as a bridge for fusion of dual vision modalities. Specifically, a paired antithetical text prompt is designed to generate binary text semantics, and triple-modal contrastive pre-training is offered to accomplish multimodal learning. Experiment results show that our proposed method achieves 73.83\% I-AUROC and 93.05\% P-AUROC with a total missing rate 0.7 for RGB and 3D modalities (exceeding state-of-the-art methods 3.84\% and 5.58\% respectively), and outperforms existing approaches to varying degrees under different missing types and rates. The source code will be available at \url{https://github.com/SvyJ/MISDD-MM}.
\end{abstract}

%% file: Chapters/1_Intro.tex
\section{Introduction}
\label{sec:intro}

Industrial surface defect detection (ISDD)~\cite{niu2021positive,wang2024robust,ma2024spdp} assures quality control in manufacturing and production systems by identifying data features that deviate from normal patterns. Due to the scarcity of negative samples, ISDD typically manifests as an unsupervised paradigm. In the past few years, researchers have devoted much effort and greatly advanced the progress of unsupervised ISDD~\cite{wang2024robust,roth2022towards,chen2024unified}. However, early research was mainly limited to the RGB modality, which was not sufficient for applying to capture defective targets with geometric features~\cite{horwitz2023back,bergmann2021mvtec}. This limitation has driven the emergence of multimodal industrial surface defect detection (MISDD), it supplements the necessary geometric priors by introducing 3D modalities such as point clouds~\cite{wang2023multimodal,chu2023shape}, depth~\cite{zavrtanik2024cheating,rudolph2023asymmetric}, or surface normals~\cite{wang2020applying}.

\begin{figure}[!ht]
\centering
\includegraphics[width=8cm]{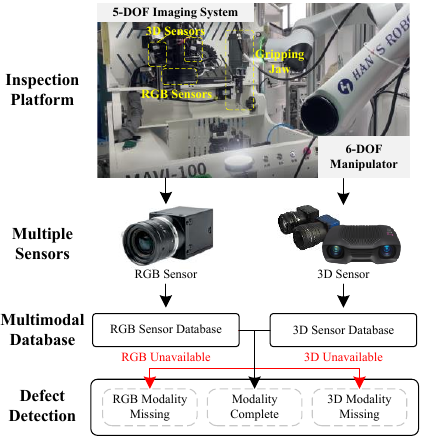}
\captionsetup{justification=justified}
\caption{Modalities missing scenarios caused by the uncertain availability of multiple sensors.}
\label{fig:intro_sensor}
\end{figure}

The transition from single-modal~\cite{roth2022towards,liu2024two,niu2021positive,tang2024ped,wang2024robust} to multimodal~\cite{wang2023multi,costanzino2024multimodal,chu2023shape} enables more accurate and comprehensive surface defect detection. But unexceptionally, multimodal learning including MISDD faces a common key issue, that is, the availability of various sensors. As shown in Fig.~\ref{fig:intro_sensor}, the automatic surface defect inspection platform~\cite{jiang2025category} we have developed includes automatic imaging and acquisition of RGB and 3D modalities. Specifically, during the training or inference process, one or some modalities may be missing due to sensor failures, environmental limitations, or cost control. In the practical dynamic industrial environment, the problem of modalities missing can be exacerbated due to unpredictable changes in sensors availability. To this end, Sui \textit{et al.}~\cite{sui2024cross} and Miao \textit{et al.}~\cite{miao2024radar} proposed the modality-incomplete industrial surface defect detection (MIISDD) task, which focuses on static data incompleteness (inherent modality missing of training and testing data), emphasizing pseudo modality imputation of modal features and representation learning. In contrast, we propose Multimodal Industrial Surface Defect Detection with Missing Modalities (MISDD-MM), which considers dynamic and sample-specific modality absence during both training and inference. MISDD-MM is defined as a practical multimodal learning scenario where the availability of RGB and 3D inputs may fluctuate unpredictably due to sensor malfunctions, environmental conditions, or adaptive acquisition strategies. This setting requires the model to perform robust defect localization and adaptive modality fusion under incomplete and uncertain input conditions, as illustrated in Fig.~\ref{fig:intro_sensor}. Therefore, the multimodal learning of MISDD-MM would be subject to two main troubles.

\begin{figure}[!t]
\centering
\includegraphics[width=8cm]{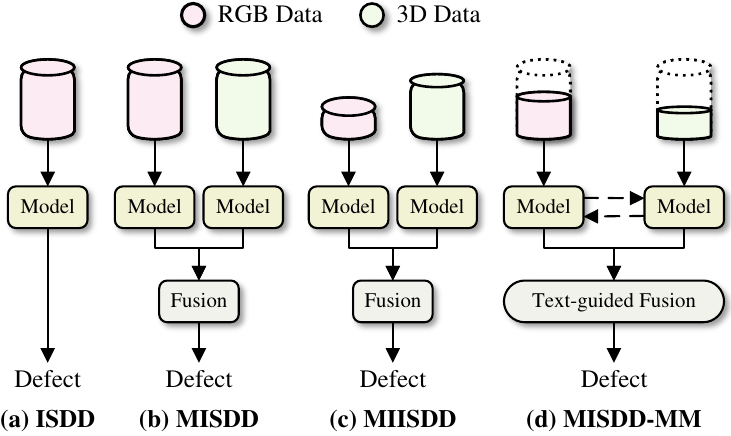}
\captionsetup{justification=justified}
\caption{Task differentiation diagram. Our proposed MISDD-MM task demands multimodal learning under dynamic missing of RGB and 3D modalities, which differs from static modality-incomplete in MIISDD.}
\label{fig:intro}
\end{figure}

Firstly, dynamic multimodal input requires the model to be able to cope with different missing types of data, which demands the ability of multimodal learning mode transformation and processing information vacancy, this is what existing multimodal models undoubtedly lack. Simply, the most direct approach is to replace or impute the missing modality, which usually leads to information distortion or modal illusion. On the contrary, it is more desirable to enable the model to adapt to the problem of missing modalities. Recently, prompt learning has been proven effective in vision-language learning~\cite{khattak2023maple,lee2023multimodal} with incomplete modalities, it leverages a few learnable parameters and optimizes them in the middle layer of the transformer through fine-tuning. But differently, the MISDD-MM task relies on dual visual modalities, requiring the model to have the ability to deal with dynamic visual inputs. To this end, we propose the cross-modal prompt learning (CPL), which includes: i) cross-modal consistency prompt for building multimodal visual information consistency for surface defect localization; ii) modality-specific prompt for adapting learn mode transformation under dynamic input patterns, to safeguard the feature extraction of a single modality. iii) missing-aware prompt for compensating information vacancy caused by dynamic modalities-missing, thus assisting cross-modal interaction. 

Secondly, dual visual modalities cannot be directly fused if one of them is missing. Alternatively, one feasible method is to map them to a shared latent space through contrastive learning. Therefore, we propose to utilize text modality to serve as a bridge for the fusion of RGB and 3D modalities. In addition, text prompts can produce both normal and abnormal patterns simultaneously~\cite{jeong2023winclip,zhu2023prompt,li2024promptad}, which would greatly benefit MISDD-MM. However, the most commonly adopted visual language model, such as CLIP~\cite{radford2021learning} and ViLT~\cite{kim2021vilt}, lacks the strength to directly process triple modalities. To cater to this special requirement, we propose symmetric contrastive learning (SCL), which first introduces antithetical text prompts to generate binary text semantics from normal-only samples. Next, the triple-modal contrastive pre-training is considered as two symmetric visual-language learning with binary prompts. 

In summary, by utilizing frozen feature extractors and in cooperation with learnable prompts in CPL and SCL, the solution to the MISDD-MM task is provided. Our contributions are summarized as follows:

\begin{itemize}
    \item We introduce a new challenging task that caters more to real-world data acquisition, \textit{i.e.,} multimodal industrial surface defect detection with missing modalities caused by uncertain sensors availability.
    \item We design a resilient MISDD-MM framework that integrates cross-modal prompt learning, enabling multimodal transformer to adapt learning mode transformation and information vacancy caused by modalities-missing.
    \item We propose the symmetric contrastive learning paradigm for triple-modal contrastive pre-training, enhancing multimodal fusion with guidance of antithetical text prompt.
    \item Our proposed method achieves state-of-the-art MISDD-MM performance under different settings of missing types and missing rates.
\end{itemize}

%% file: Chapters/2_Rel.tex
\section{Related Work}
\label{sec:related}

\subsection{Multimodal Industrial Surface Defect Detection} 
Currently, the mainstream MISDD methods leverage two modalities as inputs, \textit{i.e.,} RGB images and point cloud data (or depth maps). According to technical details, state-of-the-art MISDD methods can be divided into three types: i) \textit{Memory bank-based} methods~\cite{wang2023multimodal,chu2023shape,sui2024cross,niu2021positive} detect defects by storing normal sample features and calculating the similarity between subsequent samples and bank. ii) \textit{Teacher-student architecture-based} methods~\cite{rudolph2023asymmetric,gu2024rethinking} force the pre-trained complex teacher model to guide lightweight student models to learn feature representations of normal samples through knowledge distillation. iii) \textit{Reconstruction-based} methods~\cite{chen2023easynet,costanzino2024multimodal,zavrtanik2024cheating} typically implement defect detection by reconstructing specific images, such as from synthetic defective samples to normal samples, from 2D to 3D images, \textit{etc}. The abovementioned methods are based on the premise of modality-complete, which is not applicable to real-world data acquisition scenarios. Therefore, this article introduces a more practical task termed multimodal industrial surface defect detection with missing modalities.

\subsection{Visual Prompt Learning}
Prompt Learning has efficiently transferred from natural language processing~\cite{brown2020language} to computer vision~\cite{wang2024learning} and multimodal learning~\cite{wang2024vilt}.
Early visual prompt learning utilized the cross-modal alignment ability of pre-trained visual-language models (VLMs) (such as CLIP~\cite{radford2021learning} and ViLT~\cite{kim2021vilt}) by optimizing text prompts~\cite{zhou2022learning} or jointly optimizing visual and text prompts~\cite{khattak2023maple}. Static prompt insertion~\cite{jia2022visual} and dynamic prompt generation~\cite{shi2025deep} represent two different types of prompt learning, the former directly changing image feature representation, while the latter adaptively adjusting prompts based on input content, enhancing the adaptability to diverse scenarios. Recently, cross-modal prompt~\cite{feng2024cp,zhai2024multi} framework have also been constructed. Intuitively, the application of prompt learning in multimodal tasks is mostly concentrated in VLMs. However, in fact, its scalability in more modalities has also been widely verified, such as video~\cite{li2023compressed}, audio~\cite{huang2023make}, 3D image modalities (such as point clouds~\cite{sun2024parameter} and depth maps~\cite{ikemura2024robust}), \textit{etc}. Similarly, this article aims to establish a cross-modal prompt mechanism between RGB and 3D modalities to support multimodal learning for industrial surface defect detection.

\subsection{Multimodal Learning with Missing Modality}
Implementing multimodal learning in situations where one or some modalities are missing has more practical application prospects, but it is also highly challenging. Straightforward solutions such as modality imputation~\cite{tran2017missing} and estimation~\cite{ma2021smil} are prone to introduce noise, while shared representation learning~\cite{wang2023multi,yao2024drfuse} directly processes incomplete data through joint embedding. Recently, prompt learning has emerged as a powerful paradigm for handling missing modalities~\cite{lee2023multimodal,qiu2023modal}, offering low-cost adaptation~\cite{shi2025deep} and cross-modal knowledge transfer advantages~\cite{miao2024radar}. Hence, this article further explores the prospects of leveraging prompt learning in the task of multimodal industrial surface defect detection with missing modalities.

%% file: Chapters/3_Met.tex
\section{Approach}
\label{sec:method}

\begin{figure*}[!t]
\centering
\includegraphics[width=\textwidth]{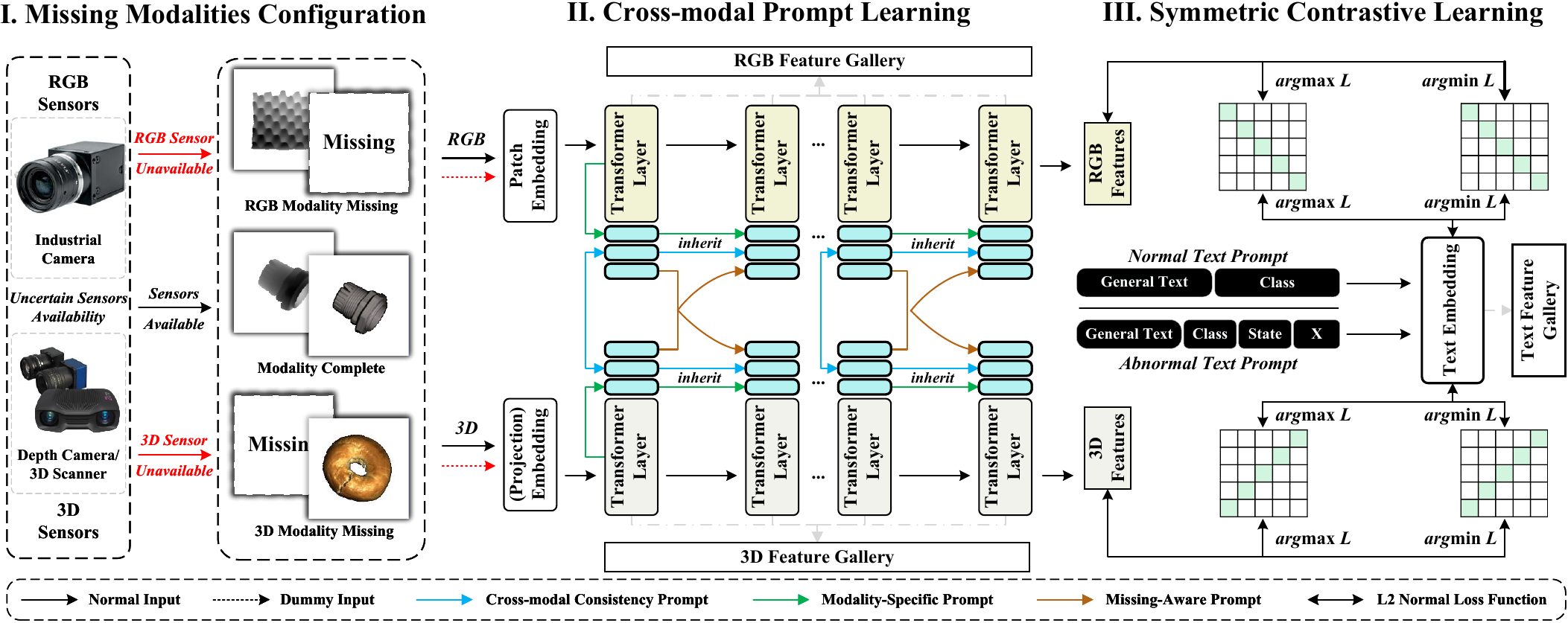}
\captionsetup{justification=justified}
\caption{The overall flowchart of our proposed framework for MISDD-MM. It consists of three serial phases: \textbf{(I)} Missing modalities configuration, which produced three input patterns through three modalities-missing settings. \textbf{(II)} Cross-modal prompt learning, which includes three specially designed prompts with colored solid lines, \textit{i.e.,} \textcolor{ccp}{cross-modal consistency prompt}, \textcolor{msp}{modality-specific prompt}, and \textcolor{map}{missing-aware prompt}. \textbf{(III)} Symmetric contrastive learning, which performs triple-modal contrastive pre-training to generate defect detection results. Prompt injection occurs at early transformer layers, where input tokens are prepended with three prompts based on current modality availability.}
\label{fig:framework}
\end{figure*}

\subsection{Overall Framework}
Fig.~\ref{fig:framework} presents the overall flowchart of our proposed method, which consists of three phases, \textit{i.e.,} missing modalities configuration, cross-modal prompt learning, and symmetric contrastive learning. Specifically, three input patterns are generated based on the missing modalities configuration, where the modality-missing data is initialized by zero padding at the input-level or feature-level, termed as dummy input. Note that as the 3D modality is either a point cloud or depth map, so there is an optional projection operation to obtain depth map from the point cloud data. Subsequently, three paired cross-modal prompts are attached to the transformer layer and are optionally inherited with the increase in depth. On the one hand, they serve as missing-aware prompts for different inputs, and on the other hand, they are utilized for adapting to feature extraction of a single modality. In addition, the galleries placed on both sides are used to store the RGB and 3D features of normal samples during the training process. Next, normal text prompts and synthetic abnormal text prompts are fed into a text tokenizer to generate semantic duality, and then perform interaction with visual features. This process is implemented through our proposed symmetric contrastive learning, which guides to narrowing the distance between normal samples and normal text prompts, while increasing that between abnormal text prompts. Similarly, a text feature gallery is established to store semantic duality for the inference process.

\subsection{Missing Modalities Configuration}
In this article, three different modalities, \textit{i.e.,} RGB, 3D, and text, are leveraged, but only RGB and 3D modalities collected from sensors are simulated to be missing due to uncertain availability of sensors. Accordingly, there are three different missing types, \textit{i.e.,} RGB missing, 3D missing, and both missing, resulting in three distinct data patterns, \textit{i.e.,} RGB-only, 3D-only, and modality-complete. Both modalities missing may appear in the dataset simultaneously, but not for each sample. To simulate such cases, the modality-missing data is replaced by dummy input, which can be initialized through the following process.

Modality missing may occur at the input level, as well as at the feature level, which we also refer to as pseudo missing. Feature-level missing is specially designed to accommodate memory bank-based methods such as M3DM~\cite{wang2023multimodal} and Shape-Guided~\cite{chu2023shape}. For the sake of unification and comparison, we have set the type and pattern of modality missing to be only dependent on the sample and not on the position of missing, and the data pattern is determined with inputs. Specifically, different data patterns are labeled with an indicating matrix $\mathcal{M}\in\mathbb{R}^{C \times H \times W}$, where $C$, $H$, and $W$ denotes the channels, height, and width of input or extracted feature maps. The definition of $\mathcal{M}$ is according to the following rules:
\begin{equation}
\label{eq:flag}
    \langle\mathcal{M}_{i}^{rgb},\mathcal{M}_{i}^{3D}\rangle = \left\{
\begin{aligned}
\langle\boldsymbol{0}, \boldsymbol{1}\rangle&,\,\text{if}\,\mathcal{D}_{i}^{rgb}\,\text{is missing}, \\
\langle\boldsymbol{1}, \boldsymbol{0}\rangle&,\,\text{if}\,\mathcal{D}_{i}^{3D}\,\text{is missing}, \\
\langle\boldsymbol{1}, \boldsymbol{1}\rangle&,\,\text{otherwise}. \\
\end{aligned}
\right.
\end{equation}

\subsubsection{Input-level Missing}
Modality missing at the input level is the main simulation method in this article, as it is more in line with actual data acquisition scenarios. Concretely, given a paired RGB image $\mathcal{D}_{i}^{rgb}$ and 3D data $\mathcal{D}_{i}^{3D}$, according to the data missing rules in Eq.~\ref{eq:flag}, the model input $\widetilde{\mathcal{X}_{i}}$ generated by the input-level missing scheme can be formulated as:
\begin{equation}
\label{eq:input_missing_x}
    \widetilde{\mathcal{X}_{i}} = \langle\mathcal{D}_{i}^{rgb}\otimes\mathcal{M}_i^{rgb}, \mathcal{D}_{i}^{3D}\otimes\mathcal{M}_i^{3D}\rangle,
\end{equation}
where $\otimes$ denotes pixel-level multiplication. Subsequently, $\widetilde{X_{i}}$ is fed into feature extractor $\Omega$ for multimodal representation, which can be expressed as:
\begin{equation}
\langle\mathcal{F}_{i}^{rgb}, \mathcal{F}_{i}^{3D}\rangle = \Omega(\widetilde{\mathcal{X}_{i}}).
\end{equation}

\subsubsection{Feature-level Missing}
Different from input-level missing, feature-level missing is designed for methods equipped with feature alignment or pre-extraction, such as memory bank and feature gallery, \textit{etc}. It is worth noting that since the feature extractor $\Omega$ is frozen, feature-level missing does not result in unfair comparisons. Therefore, it is considered as pseudo modalities missing scheme, as the input dataloder remains complete, which can be formulated as:
\begin{equation}
\label{eq:feature_missing_x}
    \mathcal{X}_{i} = \langle\mathcal{D}_{i}^{rgb}, \mathcal{D}_{i}^{3D}\rangle.
\end{equation}

Similarly, the feature extraction of modality-complete input $\mathcal{X}_{i}$, defined as:
\begin{equation}
\langle\mathcal{F}_{i}^{rgb}, \mathcal{F}_{i}^{3D}\rangle = \Omega(\mathcal{X}_{i}),
\end{equation}
then missing operation is performed on the extracted feature maps $\mathcal{F}_{i}^{rgb}$ and $\mathcal{F}_{i}^{3D}$ as the following formulation:
\begin{equation}
\langle\widetilde{\mathcal{F}_{i}^{rgb}}, \widetilde{\mathcal{F}_{i}^{3D}}\rangle = \langle\mathcal{F}_{i}^{rgb}\otimes\mathcal{M}_i^{rgb}, \mathcal{F}_{i}^{3D}\otimes\mathcal{M}_i^{3D}\rangle.
\end{equation}

\subsection{Cross-modal Prompt Learning}

\subsubsection{RGB \& 3D Feature Representation}
Vision transformer (ViT)~\cite{dosovitskiy2021vit} has strong global modeling capabilities and compatibility with text transformers, making it an ideal choice for CLIP~\cite{radford2021learning} to achieve efficient cross-modal alignment. However,~\cite{zhou2022extract,li2024promptad} disclosed the fact that global feature representation based on $Q$-$K$ self-attention actually weakens the localization ability of CLIP. Instead, Li \textit{et al.}~\cite{li2025closer} proposed a consistent self-attention to build local relations for consistent semantics, which relies on the self-attention matrix $VV^T$. The implementation is as follows:
\begin{equation}
\label{eq:self-attention}
    \text{Attention}(V,V,V) = \text{softmax}(\frac{VV^T}{\sqrt{d_k}})V.
\end{equation}

In this article, ViT based on consistent self-attention is leveraged to extract features from RGB images and depth maps, while PointMAE~\cite{pang2022masked} is adopted for point cloud data. In addition, two feature galleries $G^{rgb}$ and $G^{3D}$ are established to store extracted features of normal training samples from RGB and 3D modalities.

\subsubsection{Missing-aware Cross-modal Prompt}
Modalities-missing brings several troubles for cross-modal learning~\cite{lee2023multimodal,shi2025deep}, including: i) Problem of establishing multimodal information consistency; ii) Learning mode transformation under different input patterns; iii) Information vacancy under modalities-missing. Correspondingly, we propose three distinct prompts to address the above issues. Note that considering that multimodal learning under missing modalities in this article is based on dual visual modalities, we assign the same prompts to the pre-trained transformer layers for RGB and 3D modalities. Below three well-designed prompts are introduced in turn.

\noindent
\textbf{Cross-modal Consistency Prompt}. As introduced in Eq.~\ref{eq:flag}, modality-missing configurations leading three different input patterns, which makes it difficult to establish and store consistent information between RGB and 3D modalities. To this end, a cross-modal consistency prompt (CCP) is proposed to ensure a consistent buffer area exists under different inputs. Simply but effectively, CCP is a randomly initialized parameter matrix $\mathcal{P}_1\in\mathbb{R}^{l \times d}$, where $l$ is the predetermined length of prompt, and $d$ represents the embedding dimensions of transformer layer. In addition, CCP is attached to the first layer in parallel feature extractors.

\noindent
\textbf{Modality-specific Prompt}. To cope with the difference resulting from modality-complete or modality-missing in the single modal branch, we design a modality-specific prompt (MSP) for RGB or 3D modality, realizing the function of updating the prompt content based on different input patterns. Specifically, MSP is generated based on model input and is inherited as the number of layers increases, which leverages self-attention mentioned in Eq.~\ref{eq:self-attention} and can be formulated as:
\begin{equation}
\label{eq:msp}
    \mathcal{P}_2 = \Pi_{i=1}^h\,\text{softmax}(\frac{\mathcal{W}^{h}\mathcal{X}^m(\mathcal{W}^{h}\mathcal{X}^m)^T)}{\sqrt{d_k}})\mathcal{W}^{h}\mathcal{X}^m,
\end{equation}
where $\Pi$ is concatenation function, while $h$ denotes the number of heads in self-attention, $\mathcal{W}^{h}$ represents projection matrix, and $\mathcal{X}^m$ is the input $\mathcal{D}^m$ of modality $m$ in Eqs.~\ref{eq:input_missing_x} or~\ref{eq:feature_missing_x}. Therefore, learning mode transformation under different input patterns can be adapted with by utilizing of MSP.

\noindent
\textbf{Missing-aware Prompt}. Information vacancy is the most serious problem caused by modalities missing, which requires each layer of the transformer to have the ability to deal with it independently~\cite{lee2023multimodal}. Based on this observation and similar to CCP and MSP, missing-aware prompt (MAP) $\mathcal{P}_3$ is designed to attach to all layers. Concretely, the size and initialization of MAP are the same as CCP, but the prompts in the two branches do not share parameters. Moreover, the update of MAP in the transformer layers is cascaded, which in $j$-th layer can be formulated as Eq.~\ref{eq:msp} by concatenating $\mathcal{P}_3$ with $\mathcal{X}^m$. The insertion of MAP compensates for the adaptability of information vacancy in cross-modal learning, avoiding feature fragmentation caused by dynamic modalities-missing.

\noindent
\textbf{Prompt Injection}. The prompts are integrated into the modified residual attention blocks of the visual encoder, specifically at the early transformer layers (\textit{i.e.}, blocks where the layer index is zero and the depth of the layer falls within the predefined prompt depth, which is set to 6 by default). Specifically, at each relevant layer, a multi-head attention operation is used to refine three prompts in relation to the available visual features, followed by sequence concatenation. The concatenated prompt-enhanced sequence is then passed through the standard attention and MLPs.

\subsection{Symmetric Contrastive Learning}

The overall design of symmetric contrastive learning adopts a dual-branch structure where two visual modalities (RGB and 3D) are independently aligned with the text modality. Each branch shares the same architecture and learning mechanism, forming two parallel and symmetric pathways for vision-language contrastive learning, as illustrated in Fig.~\ref{fig:framework}.

\subsubsection{Antithetical Text Prompt}
Positive sample-based prompt learning cannot achieve effective contrastive learning, so we build negative sample prompts for the text modality. Inspired by~\cite{zhou2022learning,li2024promptad}, the semantic inversion of text can be achieved by concatenating abnormal ``STATE''. As shown in Fig.~\ref{fig:framework}, normal and abnormal text prompts with binary semantics are constructed in a structurally symmetric way. Specifically, the normal text prompt that fed into the text tokenizer is designed as $\mathcal{T}^{n} = [\text{General Text}][\text{Class}]$, where $[\text{General Text}]$ represents sentence like ``A photo of a ...'', which is the same for all defect classes. On the contrary, the abnormal text prompt is implemented by affixing a ``State'' to a normal text prompt, such as ``with crack''. In addition, to ensure its flexibility, we have respectively set fixed and learnable suffixes $[X]$ for abnormal text prompts, which are designed as $\mathcal{T}^{an} = [\text{General Text}][\text{Class}][\text{State}][\text{X}]$. Based on this design, symmetric input patterns are established for the contrastive learning objectives.

\subsubsection{Triple-modal Contrastive Pre-training}
Based on the binary semantic embedding $T^{n}$ and $T^{an}$ generated by the antithetical text prompt, while RGB features $F^{rgb}$ and 3D features $F^{3D}$ under modality missing are obtained through cross-modal prompt learning. As can be seen from Fig.~\ref{fig:framework}, due to the involvement of features of three modalities, the optimization is considered as two sets of symmetrical vision-language contrastive learning. The contrastive loss is designed as the sum of two sub-losses, each operating between a visual modality and the binary text embeddings. Specifically, the parameters $\Theta$ of cross-modal prompts and text prompts are optimized through:
\begin{align}
    \nonumber\Theta =\, &\arg\min[L(F^{rgb}, T^n) + L(F^{3D}, T^n) - \\
    &\quad\quad\quad\,\,\,\, L(F^{rgb}, T^{an}) - L(F^{3D}, T^{an})],
\end{align}
where $L$ is defined as the L2 norm of $F^m$ and $T^t$, where $m$ denotes RGB or 3D modality, and $t$ denotes normal or abnormal text embedding.

\subsection{Multimodal Defect Detection}
All prompts are fine-tuned through cross-modal prompt learning and triple-modal contrastive pre-training, with intermediate visual and text features stored in $G^{rgb}$, $G^{3D}$, and $G^{text}$. Based on these, the partial image-level score $I$ and pixel-level map $P$ are obtained as:
\begin{align}
    I^{m_1,m_2}_i &= F_i^{m_1}\cdot(G_i^{m_2})^\mathrm{T}, \\
    P^{m_1,m_2}_i &= \sum_{i=1}^k(1-F_i^{m_1}\cdot(G_i^{m_2})^\mathrm{T}),
\end{align}
where $m_1$ and $m_2$ represent different modalities, and $k$ denotes the number of layers that generated middle features. Consequently, the final detection scores $S^{im}$ are formulated as:
\begin{equation}
    S_i^{im} = \max_i\{I^{rgb,text}_i,I^{3D,text}_i\}, 
\end{equation}
and the segmentation maps $S^{px}$ are calculated as:
\begin{equation}
    S_i^{px} = \max_i\{H(I^{rgb,text}_i,P^{rgb,text}_i),H(I^{3D,text}_i,P^{3D,text}_i)\},
\end{equation}
where $H(\cdot)$ denotes harmonic mean operation~\cite{jeong2023winclip}, which assign greater weight to normality prediction.

%% file: Chapters/4_Exps.tex
\begin{table*}[!ht]
\centering
\caption{Performance comparison of I-AUROC (\%), P-AUROC (\%), and AUPRO (\%) scores on MVTec 3D-AD dataset under different missing modality / rate ($\eta$). Missing modality = ``both'' represents RGB images and 3D data are missing with the rate of $\frac{\eta}{2}$ respectively.}
\label{tab:res}
\setlength{\tabcolsep}{0.7mm}
\begin{tabular}{ccccccccccc}
\hline
\multirow{3}{*}{\textbf{\makecell{Missing \\ Modality}}} & 
\multirow{3}{*}{\textbf{Method}} & 
\multicolumn{9}{c}{\textbf{Missing Rate}} \\
\cline{3-11}
 &  & 
\multicolumn{3}{c}{\textbf{$\eta$ = 30\%}} & 
\multicolumn{3}{c}{\textbf{$\eta$ = 50\%}} & 
\multicolumn{3}{c}{\textbf{$\eta$ = 70\%}} \\
\cline{3-11}
 &  & 
\textbf{I-AUROC} & \textbf{P-AUROC} & \textbf{AUPRO} & 
\textbf{I-AUROC} & \textbf{P-AUROC} & \textbf{AUPRO} & 
\textbf{I-AUROC} & \textbf{P-AUROC} & \textbf{AUPRO} \\
\hline
\multirow{6}{*}{RGB}
 & M3DM \textsubscript{2023}~\cite{wang2023multimodal} & 76.72 & 83.65 & 72.15 & 71.02 & 89.87 & 71.59 & 74.52 & \textbf{92.32} & 77.12  \\
 & Shape-Guided \textsubscript{2023}~\cite{chu2023shape} & 77.51 & 88.38 & 69.65 & 71.52 & 87.40 & 61.06 & \textbf{78.02} & 88.80 & 64.22  \\
 & AST \textsubscript{2023}~\cite{rudolph2023asymmetric} & 72.37 & 78.93 & 67.08 & 65.97 & 75.52 & 54.76 & 58.27 & 77.27 & 42.21  \\
 & CFM \textsubscript{2024}~\cite{costanzino2024multimodal} & 74.13 & 88.45 & 77.90 & \textbf{74.63} & 88.21 & \textbf{78.31} & 71.47 & 88.95 & \textbf{77.44}  \\
 & 3DSR \textsubscript{2024}~\cite{zavrtanik2024cheating} & 52.46 & 55.80 & 21.30 & 48.78 & 56.89 & 20.91 & 50.90 & 42.50 & 12.39  \\
 & \textbf{Ours} & \textbf{78.80} & \textbf{93.38} & \textbf{79.20} & 73.86 & \textbf{90.78} & 70.50 & 68.75 & 89.22 & 65.31 \\
\hline
\multirow{6}{*}{3D}
 & M3DM \textsubscript{2023}~\cite{wang2023multimodal} & 70.41 & 78.92 & 29.37 & 68.89 & 73.70 & 47.10 & 68.60 & 77.07 & 65.07  \\
 & Shape-Guided \textsubscript{2023}~\cite{chu2023shape} & 75.31 & 87.77 & 59.35 & 64.84 & 83.51 & 48.29 & 56.09 & 82.65 & 48.27  \\
 & AST \textsubscript{2023}~\cite{rudolph2023asymmetric} & 76.83 & 93.78 & 83.41 & 75.10 & 92.51 & 79.85 & 73.99 & 92.26 & 78.02   \\
 & CFM \textsubscript{2024}~\cite{costanzino2024multimodal} & 70.76 & 87.49 & 79.36 & 65.44 & 87.17 & 78.37 & 57.03 & 85.71 & 75.18  \\
 & 3DSR \textsubscript{2024}~\cite{zavrtanik2024cheating} & 50.25 & 55.48 & 25.59 & 54.33 & 55.90 & 27.35 & 47.36 & 58.01 & 27.45  \\
 & \textbf{Ours} & \textbf{81.97} & \textbf{98.16} & \textbf{92.68} & \textbf{81.74} & \textbf{98.10} & \textbf{92.44} & \textbf{81.82} & \textbf{97.98} & \textbf{91.99} \\
\hline
\multirow{6}{*}{Both}
 & M3DM \textsubscript{2023}~\cite{wang2023multimodal} & 73.03 & 79.00 & 49.42 & 67.72 & 72.89 & 32.33 & 62.39 & 69.77 & 33.99 \\
 & Shape-Guided \textsubscript{2023}~\cite{chu2023shape} & 75.31 & 91.70 & 73.39 & 62.84 & 88.38 & 61.28 & 54.89 & 85.36 & 53.08 \\
 & AST \textsubscript{2023}~\cite{rudolph2023asymmetric} & 71.87 & 84.33 & 73.20 & 70.74 & 79.72 & 67.89 & 65.71 & 79.96 & 63.78 \\
 & CFM \textsubscript{2024}~\cite{costanzino2024multimodal} & 74.22 & 89.73 & 79.39 & 70.46 & 88.42 & 77.96 & 69.99 & 87.47 & \textbf{77.95} \\
 & 3DSR \textsubscript{2024}~\cite{zavrtanik2024cheating} & 55.57 & 57.00 & 23.26 & 59.22 & 52.25 & 23.15 & 53.15 & 44.28 & 16.78 \\
 & \textbf{Ours} & \textbf{77.71} & \textbf{95.00} & \textbf{84.03} & \textbf{76.95} & \textbf{93.28} & \textbf{79.79} & \textbf{73.83} & \textbf{93.05} & 77.44  \\
\hline
\end{tabular}
\end{table*}

\section{Experiments}
\label{sec:exp}

\subsection{Experiments Setup}
\subsubsection{Datasets} 
In this article, all experiments are performed on the MVTec 3D-AD~\cite{bergmann2021mvtec} dataset, which includes 2656 training samples and 1197 test samples from 10 categories collected using RGB and 3D sensors. Each sample contains a pair of RGB images and 3D data (\textit{i.e.,} colored point cloud (PC)), and depth maps are generated and aligned from the corresponding PC.

\begin{table}[!ht]
\centering
\small
\caption{Detailed implementation configuration of missing modality for different models.}
\label{tab:config}
\setlength{\tabcolsep}{1.2mm}
\begin{tabular}{cccc}
\hline
\textbf{Method} & \textbf{\makecell{2D}} & \textbf{\makecell{3D}} & \textbf{\makecell{Missing Configuration}} \\
\hline
M3DM~\cite{wang2023multimodal} & RGB & Point Cloud & Feature-level \\
Shape-Guided~\cite{chu2023shape} & RGB & Point Cloud & Feature-level \\
AST~\cite{rudolph2023asymmetric} & RGB & Depth Map & Input-level\\
CFM~\cite{costanzino2024multimodal} & RGB & Point Cloud & Feature-level \\
3DSR~\cite{zavrtanik2024cheating} & RGB & Depth Map & Input-level \\
\textbf{Ours} & RGB & Depth Map & Input-level \\
 \hline
\end{tabular}
\end{table}

\subsubsection{Implementation Details} 
All experiments are conducted on an NVIDIA GeForce RTX 4090 GPU. The proposed method is built on ViT-B/16-based~\cite{dosovitskiy2021vit} CLIP~\cite{radford2021learning} pre-trained on LAION-400M. Multimodal configurations are listed in Table~\ref{tab:config}, baseline methods including M3DM~\cite{wang2023multimodal}, Shape-Guided~\cite{chu2023shape}, and CFM~\cite{costanzino2024multimodal} adopt RGB images and PCs as input, while AST~\cite{radford2021learning}, 3DSR~\cite{zavrtanik2024cheating}, and our proposed method adopt RGB images and depth maps. Feature-level or input-level zero filling is performed on missing modality with the missing rate of $\eta$, while that of $\frac{\eta}{2}$ for each modality in missing ``both'' scenario. For model training, the image size is set to 240, and the model is optimized using SGD with a learning rate of 0.02, momentum of 0.9, and weight decay of 0.0005. To stabilize training and improve convergence, a cosine annealing learning rate scheduler is applied, where the learning rate gradually decreases from 0.02 to a minimum of 1e-5 over the full number of training epochs. Except for the special descriptions mentioned above, other parameter configurations can refer to WinCLIP~\cite{jeong2023winclip} and M3DM~\cite{wang2023multimodal}.

\begin{table*}[!ht]
\centering
\caption{P-AUROC (\%) score for defect detection of all categories of MVTec 3D-AD dataset with both modality missing.}
\label{table:p-auroc_both_main}
\setlength{\tabcolsep}{1mm}
\begin{tabular}{ccccccccccccc}
\hline
\textbf{\makecell{Missing Modality /\\ Missing Rate}} & \textbf{Methods} & \multicolumn{1}{c}{\textbf{Bagel}} & \multicolumn{1}{c}{\textbf{\makecell{Cable \\Gland}}} & \multicolumn{1}{c}{\textbf{Carrot}} & \multicolumn{1}{c}{\textbf{Cookie}} & \multicolumn{1}{c}{\textbf{Dowel}} & \multicolumn{1}{c}{\textbf{Foam}} & \multicolumn{1}{c}{\textbf{Peach}} & \multicolumn{1}{c}{\textbf{Potato}} & \multicolumn{1}{c}{\textbf{Rope}} & \multicolumn{1}{c}{\textbf{Tire}} & \multicolumn{1}{c}{\textbf{Mean}} \\
\hline
\multirow{6}{*}{\makecell{RGB / 15\%\\ 3D / 15\%}} 
 & M3DM \textsubscript{2023}~\cite{wang2023multimodal} & 74.52 & 77.77 & 85.02 & 76.02 & 78.49 & 79.36 & 77.78 & 79.36 & 82.79 & 78.84 & 79.00 \\
 & Shape-Guided \textsubscript{2023}~\cite{chu2023shape} & 88.92 & 91.87 & 95.67 & 85.78 & 92.77 & 86.05 & 92.57 & \textbf{95.37} & 95.02 & 93.02 & 91.70 \\
 & AST \textsubscript{2023}~\cite{rudolph2023asymmetric} & 86.12 & 85.87 & 87.08 & 92.48 & 86.44 & 75.28 & 86.64 & 83.63 & 82.75 & 77.02 & 84.33 \\
 & CFM \textsubscript{2024}~\cite{costanzino2024multimodal} & 90.83 & 87.92 & 91.77 & 84.19 & 92.56 & \textbf{89.82} & 87.82 & 93.41 & 91.30 & 87.67 & 89.73 \\
 & 3DSR \textsubscript{2024}~\cite{zavrtanik2024cheating} & 70.04 & 67.96 & 63.45 & 62.36 & 59.92 & 74.08 & 37.44 & 39.52 & 28.14 & 67.07 & 57.00 \\
 & \textbf{Ours} & \textbf{94.60} & \textbf{95.00} & \textbf{97.84} & \textbf{94.79} & \textbf{97.54} & 88.64 & \textbf{93.45} & 94.54 & 9\textbf{9.04} & \textbf{94.51} & \textbf{95.00} \\
\hline
\multirow{6}{*}{\makecell{RGB / 25\%\\ 3D / 25\%}} 
 & M3DM \textsubscript{2023}~\cite{wang2023multimodal} & 73.55 & 72.48 & 79.10 & 69.90 & 71.01 & 71.51 & 68.61 & 74.31 & 76.65 & 71.79 & 72.89 \\
 & Shape-Guided \textsubscript{2023}~\cite{chu2023shape} & 85.10 & 88.63 & 93.26 & 82.08 & 90.73 & 80.87 & 88.83 & \textbf{92.62} & 92.60 & 89.09 & 88.38 \\
 & AST \textsubscript{2023}~\cite{rudolph2023asymmetric} & 78.75 & 76.13 & 85.93 & 83.67 & 82.98 & 76.04 & 77.03 & 77.05 & 87.81 & 71.83 & 79.72 \\
 & CFM \textsubscript{2024}~\cite{costanzino2024multimodal} & 91.88 & 86.31 & 88.48 & 85.08 & 89.66 & \textbf{90.06} & 88.65 & 88.00 & 86.96 & 89.14 & 88.42 \\
 & 3DSR \textsubscript{2024}~\cite{zavrtanik2024cheating} & 29.27 & 56.97 & 65.75 & 39.10 & 48.61 & 42.34 & 39.71 & 67.61 & 70.14 & 62.95 & 52.25 \\
 & \textbf{Ours} & \textbf{92.32} & \textbf{92.97} & \textbf{97.28} & \textbf{93.99} & \textbf{97.26} & 87.29 & \textbf{90.57} & 90.78 & \textbf{99.05} & \textbf{91.25} & \textbf{93.28} \\
\hline
\multirow{6}{*}{\makecell{RGB / 35\%\\ 3D / 35\%}} 
 & M3DM \textsubscript{2023}~\cite{wang2023multimodal} & 66.36 & 69.40 & 73.88 & 70.62 & 68.16 & 67.48 & 68.30 & 70.10 & 75.62 & 67.79 & 69.77 \\
 & Shape-Guided \textsubscript{2023}~\cite{chu2023shape} & 76.65 & 86.22 & 91.02 & 79.32 & 89.30 & 76.83 & 86.01 & 90.71 & 91.44 & 86.06 & 85.36 \\
 & AST \textsubscript{2023}~\cite{rudolph2023asymmetric} & 77.16 & 75.33 & 79.78 & 83.40 & 83.03 & 72.25 & 82.39 & 86.89 & 86.72 & 72.66 & 79.96 \\
 & CFM \textsubscript{2024}~\cite{costanzino2024multimodal} & 86.60 & 87.65 & 90.61 & 83.47 & 88.42 & 83.63 & 88.09 & 90.44 & 89.60 & 86.17 & 87.47 \\
 & 3DSR \textsubscript{2024}~\cite{zavrtanik2024cheating} & 33.75 & 39.86 & 43.21 & 45.58 & 42.29 & 39.10 & 46.92 & 43.60 & 51.62 & 56.82 & 44.28 \\
 & \textbf{Ours} & \textbf{93.36} & \textbf{90.88} & \textbf{96.40} & \textbf{94.31} & \textbf{96.21} & \textbf{85.75} & \textbf{91.31} & \textbf{93.33} & \textbf{99.14} & \textbf{89.80} & \textbf{93.05} \\
\hline
\end{tabular}
\end{table*}

\subsubsection{Evaluation Metrics} 
Following the commonly adopted evaluation metrics, image-level defect detection performance is evaluated with image-level area under the receiver operator curve (I-AUROC), while pixel-level defect segmentation performance is evaluated with pixel-level area under the receiver operator curve (P-AUROC) and area under the per-region overlap (AUPRO), which adopts 0.3 as the false positive rate integration threshold. Specifically, given image inputs $\mathcal{D}_i^{rgb}$, $\mathcal{D}_i^{3D}$ and the corresponding true anomaly label $\mathcal{Y}_i\in\{0,1\}$, the prediction score $S_i$ (regardless of image-level score $S_i^{im}$ or pixel-level score $S_i^{px}$) is generated by processing multimodal inputs with uncertain missing, the AUROC score is defined as:
\begin{align}
\text{AUROC} &= \int_{0}^{1} \text{TPR}(f)\cdot d\text{FPR}(f), \tag{14}\\
\text{TPR}(f) &= \frac{\sum_i \mathbb{I}(s_i\geq f \land y_i=1)}{\sum_i \mathbb{I}(y_i=1)}, \tag{15}\\
\text{FPR}(f) &= \frac{\sum_i \mathbb{I}(s_i\geq f \land y_i=0)}{\sum_i \mathbb{I}(y_i=0)} \tag{16},
\end{align}
where $\mathbb{I}$ is indicator function, and $f$ denotes candidate threshold, which derives from the ascending sort of scores in $s_i$. In addition, the AUPRO score is defined as:
\begin{align}
\text{AUPRO} &= \int_{0}^{1} \text{PRO}(f)\cdot df, \tag{17}\\
\text{PRO}(f) &= \frac{1}{N}\sum_{k=1}^N \mathbb{I}(\text{IoU}_k (f)\geq \tau),(\tau=0.3), \tag{18}\\
\text{IoU}_k (f) &= \frac{|R_k \cap P_k(f)|}{|R_k \cup P_k(f)|}, \tag{19}
\end{align}
where $N$ is the number of anomalous areas, $R_k$ and $P_k$ denote the binary map of ground truth and prediction results, and $\tau$ is the IoU threshold.

\subsection{Main Results}
\subsubsection{Quantitative Comparison}
Table~\ref{tab:res} presents the experimental results of the MISDD-MM task on the MVTec 3D-AD dataset~\cite{bergmann2021mvtec}, the I-AUROC, P-AUROC and, and AUPRO scores under different types and rates of modalities-missing are reported for comparison. Specifically, if the missing type is set to ``RGB'' or ``3D'', it means that the RGB image or 3D data (point cloud or depth map) is randomly missing at a rate $\eta$. while both RGB images and 3D data are missing with $\frac{\eta}{2}$ if the missing type is set to ``both''. It can be seen from Table~\ref{tab:res} that our proposed method has achieved state-of-the-art performance in nearly all missing settings, except for the case where 70\% of RGB images are missing. Under this setting, MISDD can almost only rely on the 3D modality, while better performing methods, such as M3DM~\cite{wang2023multimodal}, Shape Guided~\cite{chu2023shape} and CFM~\cite{costanzino2024multimodal}, utilize color point cloud data as the 3D modality input, which can provide a small amount of color information to compensate for the vacancy of RGB modality.

\begin{figure}[!t]
\centering
\includegraphics[width=\columnwidth]{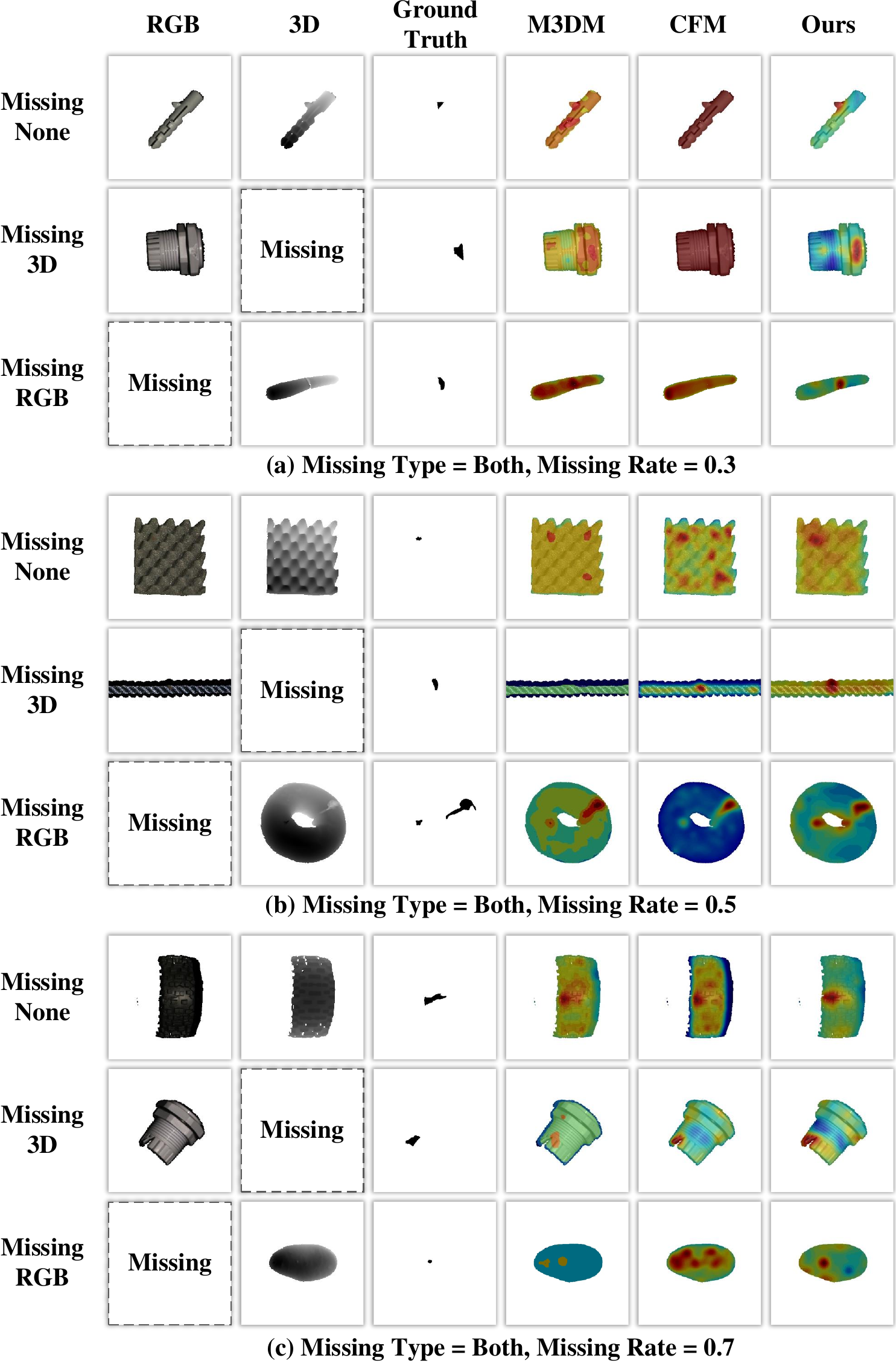}
\captionsetup{justification=justified}
\caption{Visual comparison of prediction produced by the model trained with both modalities missing under different rates.}
\label{fig:com-both}
\end{figure}

Additionally, defect segmentation results of all categories with ``both'' modalities missing are reported in Table ~\ref{table:p-auroc_both_main}. It can be seen that our proposed method achieves 95.00\%, 93.28\%, and 93.05\% on the P-AUROC scores under different missing rates while RGB and 3D modalities are missing simultaneously. Even with a missing rate of up to 70\%, it can still achieve a P-AUROC score of over 93\%, which demonstrates the effectiveness of the proposed method. Compared to other state-of-the-art MISDD methods, 3.3\%, 4.86\%, and 5.58\% leading the runner-ups also show its progressiveness.

\subsubsection{Visualization Analysis}
Fig.~\ref{fig:com-both} presents the visual comparison of defect detection results. It can be seen from Fig.~\ref{fig:com-both} (b) and (c), our proposed method can still effectively detect defects under high missing rates. In addition, we can also observe that methods such as M3DM and CFM have a significant dependence on RGB data, as they have inaccurate inference results when only 3D data are available. Benefiting from CPL, cross-modal compensation enables our proposed method to adapt to different modalities-missing situations. Due to space limitations, we only analyze the detection performance under different missing rates in the case of both modalities missing.

\subsection{Ablation Study}
\begin{figure*}[!ht]%
\centering
\subfloat[Both modalities missing.]{
    \includegraphics[width=0.33\linewidth]{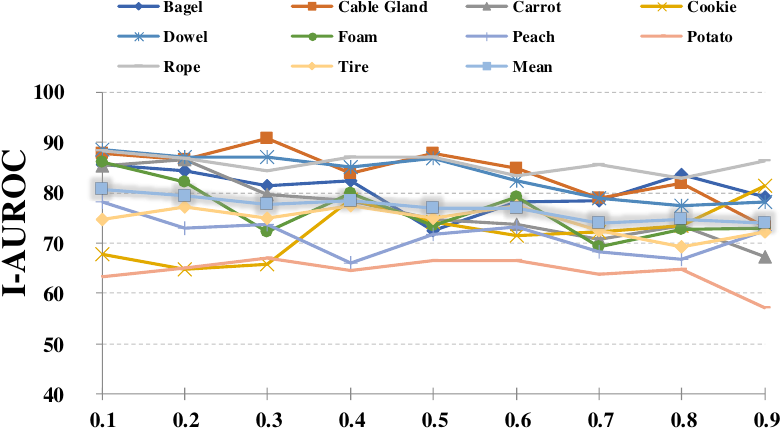}
    }
\subfloat[RGB modality missing.]{
    \includegraphics[width=0.33\linewidth]{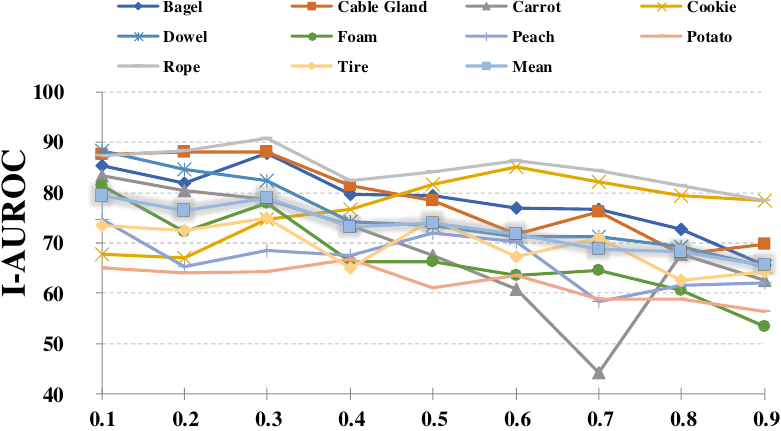}
    }
\subfloat[3D modality missing.]{
    \includegraphics[width=0.33\linewidth]{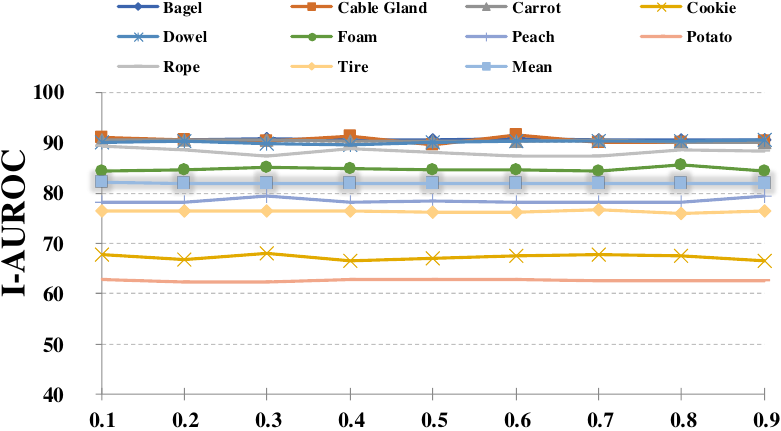}
    }
\captionsetup{justification=justified}
\caption{Variation trend of I-AUROC scores for all categories under different missing types and rates.}
\label{fig:i-auroc_trend}
\end{figure*}

\begin{figure}[!t]
\centering
\includegraphics[width=\columnwidth]{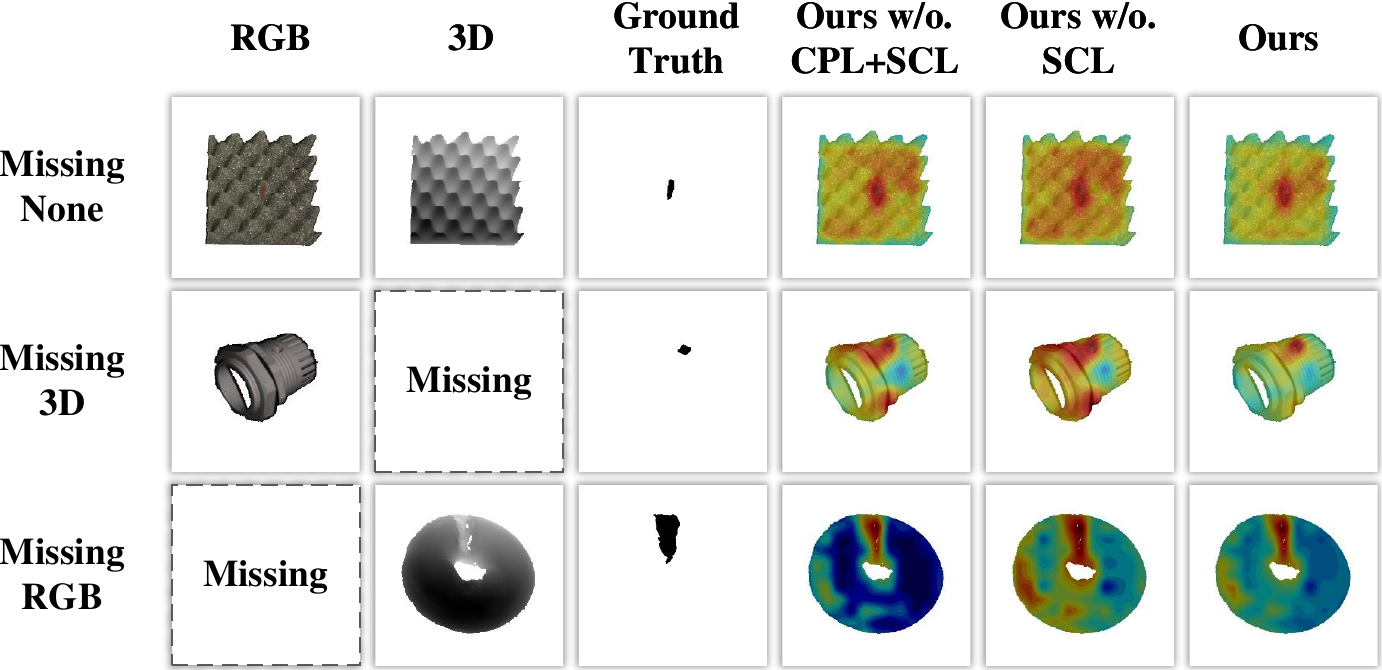}
\captionsetup{justification=justified}
\caption{Example ablation study on model trained with both modalities missing and missing rate of 70\%. The last three columns show the visualization effect of the available modality (for inference) superposition prediction results.}
\label{fig:ablation}
\end{figure}

We conduct ablation analysis on the two main pivots proposed in this article, \textit{i.e.,} CPL and SCL. The quantitative results with increments offered by CPL and SCL are reported in Table~\ref{tab:ablation}). Besides, visualization results of ablation studies are provided in Fig.~\ref{fig:ablation}. The impact analysis of the missing rate has also been reported in Fig.~\ref{fig:i-auroc_trend}.

\begin{table*}[!ht]
\centering
\caption{Ablation study of proposed cross-modal prompt learning and symmetric contrastive learning. $\Delta_1$ indicates the increment relative to the baseline (first row under each setting) after adopting CPL, while $\Delta_2$ denotes that after adopting a combination of CPL and SCL.
}
\label{tab:ablation}
\setlength{\tabcolsep}{1.2mm}
\begin{tabular}{ccccccccccc}
\hline
\multirow{3}{*}{\textbf{\makecell{Missing \\ Modality}}} & 
\multirow{3}{*}{\textbf{Method}} & 
\multicolumn{9}{c}{\textbf{Missing Rate}} \\
\cline{3-11}
 &  & 
\multicolumn{3}{c}{\textbf{$\eta$ = 30\%}} & 
\multicolumn{3}{c}{\textbf{$\eta$ = 50\%}} & 
\multicolumn{3}{c}{\textbf{$\eta$ = 70\%}} \\
\cline{3-11}
 &  & 
\textbf{I-AUROC} & \textbf{P-AUROC} & \textbf{AUPRO} & 
\textbf{I-AUROC} & \textbf{P-AUROC} & \textbf{AUPRO} & 
\textbf{I-AUROC} & \textbf{P-AUROC} & \textbf{AUPRO} \\
\hline
\multirow{5}{*}{RGB}
 & Ours w/o. [CPL+SCL] & 78.87 & 85.40 & 71.29 & 69.75 & 81.88 & 57.72 & 60.36 & 82.37 & 47.17  \\
 & Ours w/o. [SCL] & \textbf{79.99} & 93.15 & 78.44 & 70.57 & \textbf{91.22} & \textbf{71.70} & 61.65 & \textbf{90.67} & \textbf{68.85}  \\
 & \cellcolor{gray!30}$\Delta_1$ & \cellcolor{gray!30}1.12 & \cellcolor{gray!30}7.75 & \cellcolor{gray!30}7.15 & \cellcolor{gray!30}0.82 & \cellcolor{gray!30}9.34 & \cellcolor{gray!30}13.98 & \cellcolor{gray!30}1.29 & \cellcolor{gray!30}8.30 & \cellcolor{gray!30}21.68 \\
 & Ours & 78.80 & \textbf{93.38} & \textbf{79.20} & \textbf{73.86} & 90.78 & 70.50 & \textbf{68.75} & 89.22 & 65.31 \\
 & \cellcolor{gray!30}$\Delta_2$ & \cellcolor{gray!30}-0.07 & \cellcolor{gray!30}7.98 & \cellcolor{gray!30}7.91 & \cellcolor{gray!30}4.11 & \cellcolor{gray!30}8.9 & \cellcolor{gray!30}12.78 & \cellcolor{gray!30}8.39 & \cellcolor{gray!30}6.85 & \cellcolor{gray!30}18.14 \\
\hline
\multirow{5}{*}{3D}
 & Ours w/o. [CPL+SCL] & 83.31 & 97.18 & 89.05 & 80.14 & 96.74 & 87.47 & 77.40 & 96.76 & 87.21  \\
 & Ours w/o. [SCL] & \textbf{84.24} & 95.87 & 84.50 & 81.28 & 95.17 & 81.94 & 78.64 & 95.33 & 82.06  \\
 & \cellcolor{gray!30}$\Delta_1$ & \cellcolor{gray!30}0.93 & \cellcolor{gray!30}-1.31 & \cellcolor{gray!30}-4.55 & \cellcolor{gray!30}1.14 & \cellcolor{gray!30}-1.57 & \cellcolor{gray!30}-5.53 & \cellcolor{gray!30}1.24 & \cellcolor{gray!30}-1.43 & \cellcolor{gray!30}-5.15 \\
 & Ours & 81.97 & \textbf{98.16} & \textbf{92.68} & \textbf{81.74} & \textbf{98.10} & \textbf{92.44} & \textbf{81.82} & \textbf{97.98} & \textbf{91.99} \\
 & \cellcolor{gray!30}$\Delta_2$ & \cellcolor{gray!30}-1.34 & \cellcolor{gray!30}0.98 & \cellcolor{gray!30}3.63 & \cellcolor{gray!30}1.6 & \cellcolor{gray!30}1.36 & \cellcolor{gray!30}4.97 & \cellcolor{gray!30}4.42 & \cellcolor{gray!30}1.22 & \cellcolor{gray!30}4.78 \\
\hline
\multirow{5}{*}{Both}
 & Ours w/o. [CPL+SCL] & 76.74 & 89.19 & 78.79 & 74.47 & 85.56 & 71.51 & 71.77 & 84.17 & 65.53  \\
 & Ours w/o. [SCL] & \textbf{78.04} & 93.67 & 80.10 & 74.66 & 92.25 & 76.07 & 72.11 & 92.26 & 74.45  \\
 & \cellcolor{gray!30}$\Delta_1$ & \cellcolor{gray!30}1.30 & \cellcolor{gray!30}4.48 & \cellcolor{gray!30}1.31 & \cellcolor{gray!30}0.19 & \cellcolor{gray!30}6.69 & \cellcolor{gray!30}4.56 & \cellcolor{gray!30}0.34 & \cellcolor{gray!30}8.09 & \cellcolor{gray!30}8.92 \\
 & Ours & 77.71 & \textbf{95.00} & \textbf{84.03} & \textbf{76.95} & \textbf{93.28} & \textbf{79.79} & \textbf{73.83} & \textbf{93.05} & \textbf{77.44}  \\
 & \cellcolor{gray!30}$\Delta_2$ & \cellcolor{gray!30}0.97 & \cellcolor{gray!30}5.81 & \cellcolor{gray!30}5.24 & \cellcolor{gray!30}2.48 & \cellcolor{gray!30}7.72 & \cellcolor{gray!30}8.28 & \cellcolor{gray!30}2.06 & \cellcolor{gray!30}8.88 & \cellcolor{gray!30}11.91 \\
\hline
\end{tabular}
\end{table*}

\begin{table}[!ht]
\centering
\caption{Ablation of three prompts in CPL on MVTec 3D-AD dataset. (Missing type = ``both'', missing rate = 0.3)}
\label{table:ablation_cpl}
\setlength{\tabcolsep}{2.7mm}
\renewcommand{\arraystretch}{1}
\begin{tabular}{cccccc}
\hline
\textbf{CCP} & \textbf{MSP} & \textbf{MAP} & \textbf{I-AUROC} & \textbf{P-AUROC} & \textbf{AUPRO} \\
\hline
- & - & - & 76.74 & 89.19 & 78.79 \\
$\checkmark$ & - & - & 77.23 & 91.46 & 79.85 \\
- & $\checkmark$ & - & 77.66 & 92.42 & 79.74 \\
- & - & $\checkmark$ & 77.68 & 92.01 & 79.84 \\
$\checkmark$ & $\checkmark$ & $\checkmark$ & \textbf{78.04} & \textbf{93.67} & \textbf{80.10} \\
\hline
\end{tabular}
\end{table}

\subsubsection{In-depth analysis of CPL}
The $\Delta_1$ value in Table~\ref{tab:ablation} shows the quantization increments by leveraging the CPL. Especially for defect segmentation, at least 7\% P-AUROC improvement is achieved under different RGB modality-missing rates, even gaining for 21.68\% AUROC score at a missing rate of 70\%. However, if only a partial 3D modality is missing, CPL fails to provide improvement, which indirectly indicates that when RGB modality is missing, 3D modality can compensate for its absence through our proposed CPL as depth information is more vital for detecting defects with geometric variations. On the contrary, the texture information provided by the RGB modality cannot effectively improve the detection of these stereoscopic defects when the 3D modality is missing. As also can be proved from the column 5 in Fig.~\ref{fig:ablation}, when the other modality is missing, the addition of CPL supplements the depth information (such as the edge of cable gland in row 2) and texture information (such as small holes on the bagel in row 3) , thereby improving the confidence of defect regions.

In addition, we conducted ablation analysis of three different prompt in CPL on the MVTec 3D-AD dataset with a missing type of ``both'' and a missing rate of 0.3. The ablation results in Table~\ref{table:ablation_cpl} demonstrate that each of the three proposed prompts (CCP, MSP, and MAP) contributes positively to the model's performance under the modality-missing condition. Notably, MSP and MAP yield larger improvements than CCP, suggesting that modality-specific and modality-aware cues play a more crucial role than simple contrastive alignment when handling dual-modality absence. The full configuration achieves the best results across all metrics, indicating that the proposed prompts are complementary and jointly beneficial for robust multimodal defect detection.

\subsubsection{Effectiveness of SCL}
Unlike CPL, the target task of SCL is inclined to defect recognition, as it includes CLIP-based symmetric contrastive pre-training, making it less specialized in defect localization. Table~\ref{tab:ablation} shows that if the RGB modality is partially missing, there are 3.29\% and 7.10\% improvements under missing rates of 50\% and 70\%, respectively. The utilization of SCL lifts the I-AUROC scores to varying degrees under other different missing settings. The visual results (last column in Fig.~\ref{fig:ablation}) also demonstrate the effectiveness of SCL, especially in removing false positives.

\subsubsection{Detection Performance vs. Missing Rate}
Fig.~\ref{fig:i-auroc_trend} shows the trend of I-AUROC scores for each category and mean value with different missing rates. Normally, when the missing type is set to both modalities missing, the I-AUROC score decreases with increasing missing rate. The modest performance drop indicates strong robustness against missing modalities. In particular, unlike the significant decline observed when RGB data is missing, the model's performance remains almost unchanged when only the 3D modality is unavailable, as reflected by the near-horizontal trend line. This result discloses that the lack of 3D modality has almost no effect on the I-AUROC score, demonstrating that most of the features used to identify defects come from the RGB modality. In contrast, the 3D modality that mainly serves as an information supplement has little effect on identifying defects.

\subsection{Computational Efficiency}
Table~\ref{tab:params} provides a comprehensive overview of the number of parameters required by the method proposed in this article. Notably, the majority of the model parameters are contained within the frozen transformer backbone, which remains unchanged during training. The learnable components introduced by our method—including any additional layers, adapters, or task-specific modules—constitute less than 2\% of the total parameter count.

\begin{table}[!ht]
\centering
\small
\caption{Parameter quantity statistics.}
\label{tab:params}
\setlength{\tabcolsep}{3mm}
\begin{tabular}{ccc}
\hline
\textbf{Entry} & \textbf{Setting} & \textbf{Params} \\
\hline
Vision Transformer & Frozen & 120.43M \\
Text Transformer & Frozen & 91.16M \\
Cross-modal Consistency Prompt & Learnable & 0.03M \\
Modality-specific Prompt & Learnable & 0.2M \\
Missing-aware Prompt & Learnable & 3.74M \\
Antithetical Text Prompt & Learnable & 0.02M \\
\hline
\end{tabular}
\end{table}

\subsection{Few-shot MISDD}
We extend the proposed method to the few-shot MISDD task, which can be approximatively seen as a more extreme case of data missing. Specifically, we set only $K$ samples available in the training set, and both RGB and 3D modalities missing for the others, while the test set remains complete. The experimental results are shown in Table~\ref{tab:few-shot}, it can be seen that our proposed method has good scalability on the few-shot MISDD task, and achieves optimal performance under the 4-shot setting.

\begin{table}[!ht]
\centering
\caption{Performance comparison on few-shot MISDD.}
\label{tab:few-shot}
\setlength{\tabcolsep}{2mm}
\begin{tabular}{ccccc}
\hline
\textbf{$K$- shot} & \textbf{Method} & \textbf{I-AUROC} & \textbf{P-AUROC} & \textbf{AUPRO} \\
\hline
\multirow{3}{*}{1 shot} 
& EasyNet~\cite{chen2023easynet} & 0.601 & 0.910 & 0.755 \\
& CLIP3D-AD~\cite{zuo2024clip3d} & 0.733 & 0.969 & 0.893 \\
& \textbf{Ours} & 0.719 & 0.967 & 0.877 \\
\hline
\multirow{3}{*}{2 shot} 
& EasyNet~\cite{chen2023easynet} & 0.581 & 0.836 & 0.493 \\
& CLIP3D-AD~\cite{zuo2024clip3d} & 0.740 & 0.970 & 0.902 \\
& \textbf{Ours} & 0.736 & 0.969 & 0.885 \\
\hline
\multirow{3}{*}{4 shot} 
& EasyNet~\cite{chen2023easynet} & 0.586 & 0.860 & 0.565  \\
& CLIP3D-AD~\cite{zuo2024clip3d} & 0.707 & 0.969 & 0.889 \\ 
& \textbf{Ours} & 0.765 & 0.972 & 0.892  \\
\hline
\end{tabular}
\end{table}

%% file: Chapters/5_Conc.tex
\section{Conclusion and Limitations}
\label{sec:conc}

In this article, we propose the task termed multimodal industrial surface defect detection with missing modalities caused by uncertain sensors availability. We then offered a framework integrated with cross-modal prompt learning and symmetric contrastive learning, which are leveraged to tackle the problems of learning mode transformation and information vacancy caused by dynamic modalities missing, and thus achieve text-guided fusion of dual visual modalities. Experiments under various missing settings verify the progressiveness of the proposed method.

While the proposed MISDD-MM framework demonstrates robust performance under dynamic modality-missing scenarios, several limitations remain. First, although the use of text prompts enables semantic alignment across modalities, the effectiveness of prompt generation is still constrained by the quality and diversity of manually designed templates. Second, the computational cost associated with maintaining multiple feature galleries and prompt modules, although relatively lightweight, may pose challenges for real-time deployment in resource-constrained settings. Future work will explore several directions to address these issues. One direction involves developing adaptive or generative prompt strategies to replace fixed prompt templates, thereby enhancing generalization to unseen categories or modalities.